\documentclass[letterpaper]{article} 
\usepackage{aaai24}  
\usepackage{times}  
\usepackage{helvet}  
\usepackage{courier}  
\usepackage[hyphens]{url}  
\usepackage{graphicx} 
\usepackage{natbib}  
\usepackage{caption} 
\usepackage{algorithm}
\usepackage{algorithmic}

\usepackage{booktabs}
\usepackage{multirow}
\usepackage{tabularray}

\usepackage{tikz}
\def\checkmark{\tikz\fill[scale=0.4](0,.35) -- (.25,0) -- (1,.7) -- (.25,.15) -- cycle;}

\usepackage{threeparttable}
\usepackage{tablefootnote}

\usepackage[hang,flushmargin]{footmisc}

\usepackage{newfloat}
\usepackage{listings}
\DeclareCaptionStyle{ruled}{labelfont=normalfont,labelsep=colon,strut=off} 
\floatstyle{ruled}
\newfloat{listing}{tb}{lst}{}
\floatname{listing}{Listing}
\nocopyright 

\title{DISC-MedLLM: Bridging General Large Language Models and \\Real-World Medical Consultation}

\author{
    Zhijie Bao\textsuperscript{\rm 1, \rm 2},
    Wei Chen\textsuperscript{\rm 1},
    Shengze Xiao\textsuperscript{\rm 1},
    Kuang Ren\textsuperscript{\rm 3},
    Jiaao Wu\textsuperscript{\rm 1},
    Cheng Zhong\textsuperscript{\rm 1}, \\
    Jiajie Peng\textsuperscript{\rm 2 *}, 
    Xuanjing Huang\textsuperscript{\rm 4},
    Zhongyu Wei\textsuperscript{\rm 1, \rm 5}\thanks{Corresponding Author}
}
\affiliations{
    \textsuperscript{\rm 1}{School of Data Science, Fudan University, China}\\
    \textsuperscript{\rm 2}{School of Computer Science, Northwestern Polytechnical University, China}\\
    \textsuperscript{\rm 3}{University of Toronto, Canada} \\  \textsuperscript{\rm 4}{School of Computer Science, Fudan University, China}\\ 
    \textsuperscript{\rm 5}Research Institute of Automatic and Complex Systems, Fudan University, China\\

    zhijiebao@mail.nwpu.edu.cn,
    szxiao23@m.fudan.edu.cn,
    kuang.ren@mail.utoronto.ca,
    zhong7414@gmail.com,
    jiajiepeng@nwpu.edu.cn,
    \{chenwei18, xjhuang, zywei\}@fudan.edu.cn
    
}

\usepackage{bibentry}

\begin{document}

\maketitle

\begin{abstract}

We propose DISC-MedLLM, a comprehensive solution that leverages Large Language Models (LLMs) to provide accurate and truthful medical response in end-to-end conversational healthcare services. To construct high-quality Supervised Fine-Tuning (SFT) datasets, we employ three strategies: utilizing medical knowledge-graphs, reconstructing real-world dialogues, and incorporating human-guided preference rephrasing. These datasets are instrumental in training DISC-MedLLM, surpassing existing medical LLMs in both single-turn and multi-turn consultation scenarios. Extensive experimental results demonstrate the effectiveness of the proposed model in bridging the gap between general language models and real-world medical consultation. Additionally, we release the constructed dataset and model weights to further contribute to research and development. Further details and resources can be found at \emph{\url{https://github.com/FudanDISC/DISC-MedLLM}}.

\end{abstract}

\section{Introduction}

The emergence of the telemedicine industry has reshaped the healthcare service, offering remote medical consultations, broadening access to professionals, and trimming medical costs~\citep{haleem2021telemedicine}. Additionally, intelligent medical systems have enriched online medical services by incorporating features like medical information extraction ~\citep{lin2019enhancing,zhang2020mie,chen-etal-2023-knse}, drug recommendation~\citep{he2018kernelized,zheng2021drug}, automated diagnosis~\citep{wei2018task,zhong2022hierarchical,chen2023dxformer}, and health question answering~\citep{he2020infusing,pal2022medmcqa}, etc. 

While progress has been made in developing intelligent healthcare systems, previous studies primarily focuses on specific tasks or diseases with limited applicability, creating a gap between experimental advancements and practical applications~\citep{yang2022smart}. To bridge this gap, there is a need of comprehensive solutions for a broader range of medical scenarios, and providing high-quality healthcare services to users in an end-to-end conversational manner. 


 




Recently, Large Language Models (LLMs)~\citep{chatgpt,touvron2023llama,wei2021finetuned,ouyang2022training} have showed impressive ability to follow human instructions and to engage in meaningful conversations. These developments have opened up new possibilities for building medical consultation systems. However, medical consultation scenarios are usually complicated and beyond the capability of LLMs from general domain. 

\begin{figure*}[t]
    \centering
    \includegraphics[width=14cm]{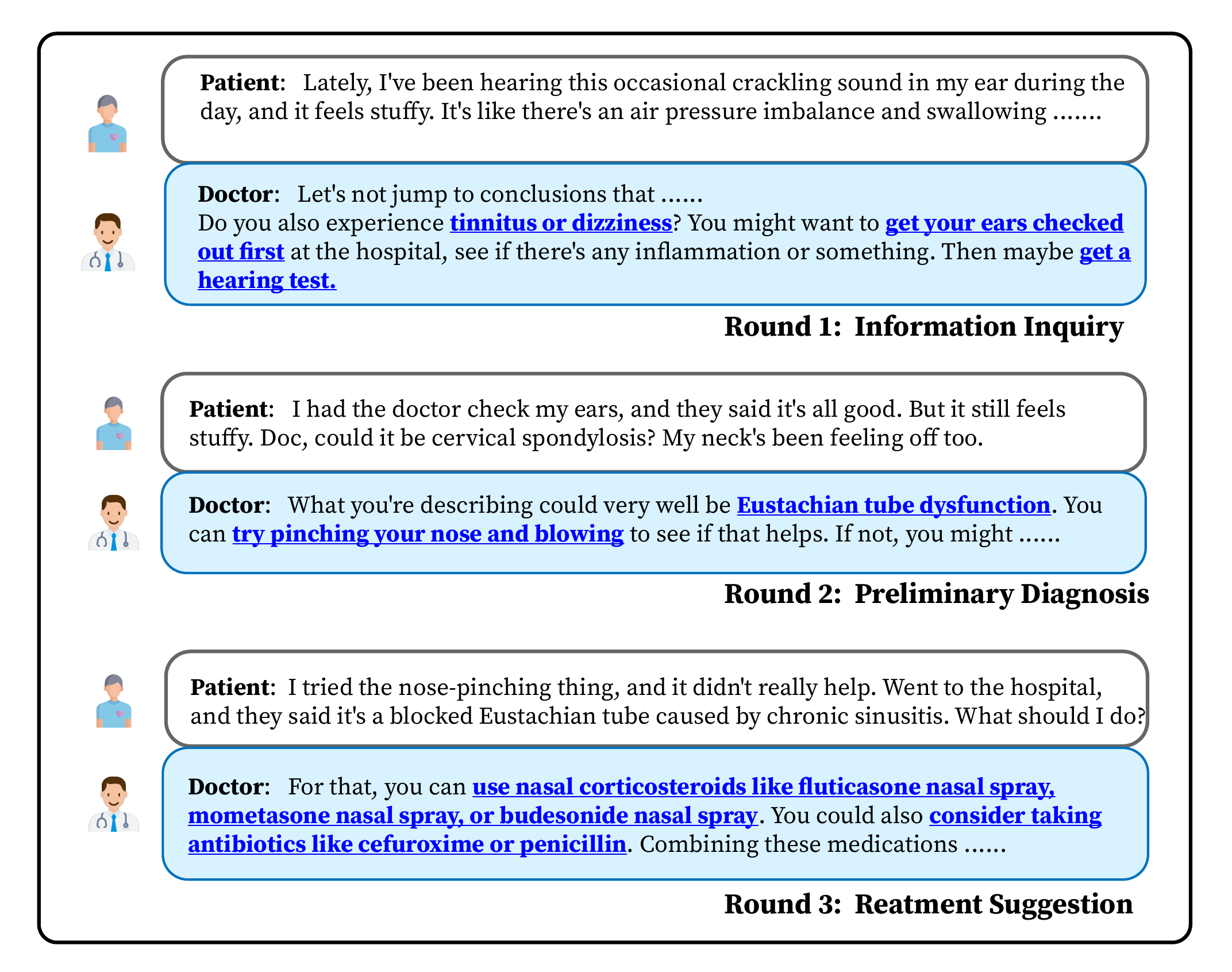}
    \caption{Dialogue Sample between a human doctor and a patient. The blue-highlighted text indicates medical entities involved in the doctor's response. Doctor's behavior reveals specific intention in each round: (1) in the round 1, further inquiries for information that aids in determining possible situations; (2) in the round 2, arrives at a preliminary diagnosis and provide valid recommendations; (3) in round 3, offers specific treatment options based on the medical condition.}
    \label{fig:real_sample}
\end{figure*}

An example of real-world medical consultation is shown in Figure~\ref{fig:real_sample}. It reveals two characteristics. Firstly, it requires intensive and reliable medical knowledge to understand the dialogue and make proper response in every steps. General domain LLMs reveal serious problems of hallucination by generating irrelevant content to the specific case. Secondly, it usually takes multiple turns to gather sufficient patient information before providing healthcare consultation and each round of conversation has specific intention. However, general domain LLMs tend to be single-turn agents with limited multi-turn inquiring capabilities regarding the details of a user's health condition. 





Based on these two observations, we argue that medical LLMs should encode intensive and reliable medical knowledge while aligning with the real-world medical dialogue distribution. Motivated by the success of Instruction Tuning~\citep{wang2022self}, we explore to construct high quality Supervised Fine Tuning (SFT) datasets for training medical LLMs, and inject medical knowledge and consultation behavior patterns into the large language model. In practice, we construct samples following three strategies. 

\begin{itemize}
    \item Medical knowledge-graph driven sample construction. We use a department-oriented strategy to sample knowledge triples from a medical knowledge graph following a patient query distribution obtained from a real-world consultation dataset. For each triple, GPT-3.5 is used to construct QA pairs in a few-shot manner. This results in 50k samples. 

    \item Real-world dialogue re-construction. Consultation records collected from medical forums are appropriate sources for LLMs fine-tuning. However, these records contain informal language usage, inconsistent terminology presentation, and different expressive style from different healthcare professionals. Therefore, we utilize GPT-3.5 to re-generate the dialogue based on real cases. This results in 420k samples.



    \item Human preference following sample collection. For the alignment of human preference, we manually select a small set of entries from the real-world medical dialogue records covering different consultation situations and manually rewrite some samples. After the human guided re-construction, we further ensure the overall quality of each dialogue. This results in 2k samples. 
\end{itemize}




The constructed SFT datasets are then utilized to train DISC-MedLLM following a two-stage training mechanism on top of a general domain Chinese LLM with 13B parameters~\footnote{In this version, we use Baichuan~\citep{Baichuan13B}as the base model. Note that our strategy can be applied to all decoder-only foundation models.}. We assess the model's performance from two perspectives to check its capability of providing accuracy answers in single-turn conversations and presenting systematical consultation in multi-turn conversations, respectively. For single-turn evaluation, we construct a benchmark consisting of multiple choices questions collected from three public medical datasets and evaluate the model's \emph{accuracy}. For multi-turn evaluation, we first construct a small set of high quality consulting cases, and then employ GPT-3.5 play the role of the patient based on the cases, and chat with the model. We use GPT-4 to evaluate the model's \emph{proactivity}, \emph{accuracy}, \emph{helpfulness} and \emph{linguistic quality}. 

The experimental results demonstrate that DISC-MedLLM outperforms the medical large-scale model HuatuoGPT~\citep{huatuogpt-2023} with same parameters (13B) by over 10\% on average in medical multiple-choice questions, although still falls behind that of GPT-3.5. Moreover, in simulated medical consultation scenarios, DISC-MedLLM exhibits superior overall performance compared to baseline models such as GPT-3.5, HuatuoGPT, and BianQue~\citep{chen2023bianque1}. Particularly in the scenarios involving medical departments and patient intents, DISC-MedLLM achieves the best performance among Chinese medical LLMs.

\begin{figure*}
\centering
\includegraphics[width=14cm]{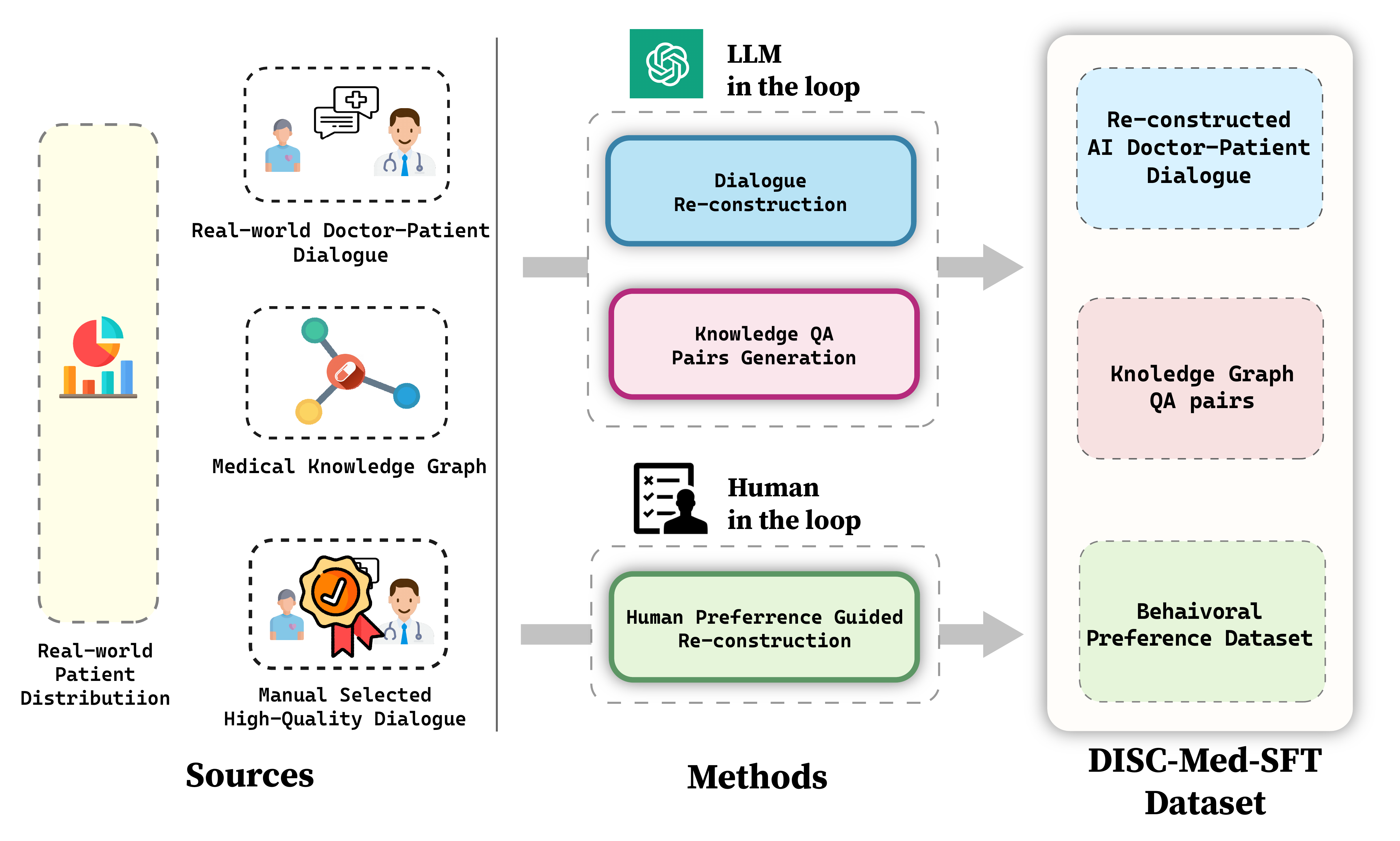}
\caption{Overview of the construction of DISC-Med-SFT. The DICS-Med-SFT dataset is constructed using various data sources, including real-world conversations and knowledge graphs, and combines the mechanisms of LLM-in-the-loop and Human-in-the-loop to form its three components: Re-constructed AI Doctor-Patient Dialogue, Knowledge Graph QA pairs, Behavioral Preference Dataset.}
\label{fig:data_construction}
\end{figure*}

\begin{table*}[t]
\centering
\begin{tabular}{@{}clrcccc@{}}
\toprule
\multirow{2}{*}{Dataset}  
& \multirow{2}{*}{Original Source} & \multirow{2}{*}{Size} & \multicolumn{4}{c}{Abilities}                                 
\\ \cmidrule(l){4-7} &  &   & 

\begin{tabular}
[c]{@{}c@{}}Domain\\ Knowledge\end{tabular} & 
\begin{tabular}[c]{@{}c@{}}Behavioral\\ Pattern\end{tabular} & 
\begin{tabular}[c]{@{}c@{}}Dialogue\\ Ability\end{tabular} & 
\begin{tabular}[c]{@{}c@{}}Human\\ Preference\end{tabular} \\ \midrule \specialrule{0em}{2pt}{2pt} 

\multirow{2}{*}{\begin{tabular}[c]{@{}c@{}}Re-constructed \\ 
AI Doctor-Patient Dialogue\end{tabular}} & MedDialog\footnotemark[2] & 400k   & \textbf{\checkmark}                              & \textbf{\checkmark}   & \textbf{\checkmark}   &       \\
& cMedQA2\footnotemark[3] & 20k   & \textbf{\checkmark}  &  &  &  \\ 
\specialrule{0em}{2pt}{2pt} 
\begin{tabular}[c]{@{}c@{}}Knowledge Graph \\
QA pairs\end{tabular}    & CMeKG\footnotemark[4] & 50k   & \textbf{\checkmark} &  &  &     \\ 
\specialrule{0em}{2pt}{2pt} 
Behavioral Preference Dataset  & Manual selection  & 2k      & & \textbf{\checkmark} &   & \textbf{\checkmark}  \\ 
\specialrule{0em}{2pt}{2pt} 
\multirow{3}{*}{Others}   & MedMCQA\footnotemark[5]   & 8k   & \textbf{\checkmark} & & &   \\ 
\specialrule{0em}{2pt}{2pt} 
   & MOSS\footnotemark[6] & 33k   &   & \textbf{\checkmark}  & \textbf{\checkmark}  \\ 
\specialrule{0em}{2pt}{2pt} 
  & Alpaca-GPT4\footnotemark[7] & 1k   &  &  &  & \textbf{\checkmark}       \\ 
\midrule                                         
\end{tabular}
\caption{Dataset Details of DISC-Med-SFT-ext, it extends from DISC-Med-SFT by incorporating general datasets. The table provides an overview of our training datasets and the corresponding capabilities they grant to the model.}
\label{tab:datasets}
\end{table*}

\section{Advantages of DISC-MedLLM}



In this section, we discuss the advantages of the proposed DISC-MedLLM, particularly its deliberated constructed dataset for fine-tuning. The overall framework is illustrated in Figure~\ref{fig:data_construction}. The dataset is primarily sourced from both medical knowledge graph and real-world doctor-patient consultations and the sample selection process is following a hospital department distribution extracted from real-world dataset. 

Two sample construction mechanisms are utilized, \textit{LLM in the loop} plays a role in paraphrasing real-world conversations, while \textit{human in the loop} ensures alignment with high-quality conversational behaviors. These efforts result in three key features of DISC-MedLLM: \emph{knowledge-intensive and reliable}, \emph{ability of multi-turn inquiry}, and \emph{alignment with human preferences}.

\subsection{Knowledge Intensive and Reliable}

In general-domain instruction tuning, it is popular to use diverse instruction examples generated by model like ChatGPT, that is, the instruction examples are derived from the model's own knowledge~\citep{wang2022self}. However, relying solely on the model itself is dangerous in the healthcare field, as LLMs itself have serious hallucination problems. Therefore, in our setting, we do not rely on the LLM to generate any medical knowledge, instead, the knowledge is entirely derived from reliable data sources, including medical knowledge graphs (from human labeled) and real doctor-patient dialogues (from doctors). The role of ChatGPT is to rewrite rather than generate. The paraphrased samples retain the underlying medical domain knowledge entirely, while ChatGPT supplements and elaborates on non-essential information. For instance, appropriate responses can provide additional details about the causes and information regarding a particular medical condition.

\subsection{Ability of Multi-turn Inquiry}


Models like ChatGPT tend to provide detailed template-like responses in one single turn when faced with healthcare inquiries, while overlooking effective inquiry and clarification of the patient's condition. In online pediatric disease consultations, for example, the average interaction between doctors and patients spans around 40 turns, with half of turns dedicated to discussing the patient's symptom details~\citep{chen2023benchmark}. DISC-MedLLM leverages real-world multi-turn patient doctor conversations, equipping the model with ability of inquiry that enable it to engage in meaningful medical inquiries. As most patients find it challenging to describe their complete physical condition in one go, possessing inquiry capabilities important and essential.




\subsection{Alignment with Human Preference}


During patient communication, human doctors exhibit concise and direct behavior. However, they often lack sufficient empathy, resulting in providing incomplete or insufficiently detailed assistance. In contrast, models like ChatGPT have behavior patterns that can be adjusted based on human preferences, leading to a tendency to provide users with as much information and help as possible. Therefore, DISC-MedLLM improves upon this by in two significant ways: 1) using human preference guided behavioral preference dataset to better align its response and behavior more closely with human preference; 2) distill the behavior patterns of ChatGPT, resulting in responses that demonstrate comprehensive explanations and a high level of empathy when engaging with patients.  \\



To a certain extent, DISC-MedLLM can be conceptualized as an amalgamation of two mentors: the first being the acquisition of medical knowledge and decision-making from doctor-patient conversations, and the second involving the assimilation of behavioral patterns and human preferences from ChatGPT. Through the integration of these two mentors and precise sampling with human intervention, DISC-MedLLM strives to align itself with the observed medical consultation distributions present in the real world.

\section{DISC-Med-SFT}
\label{sec:dataset}



To train DISC-MedLLM, we construct a high-quality dataset called DISC-Med-SFT consisting of over 470k examples derived from existing medical datasets. This comprehensive dataset encompasses various scenarios, including single-turn medical Q\&A, multi-turn medical consultations and medical multiple-choice Q\&A. Additionally, we incorporated over 34k general domain conversation and instruction samples. Detailed information regarding the employed datasets is provided in Table~\ref{tab:datasets}. 

It is worth noting that our approach differs from simply gathering a large volume of NLP datasets in the medical domain and manually creating diverse instructions as~\citeauthor{wei2021finetuned}. Instead, we adopt a goal-oriented strategy by selectively reconstructing the dataset using a few deliberately chosen sources. These data sources serve the purpose of assisting LLMs in acquiring medical domain knowledge, aligning behavioral patterns with human preferences, and capturing real-world online medical dialogue distributions.



\footnotetext[2]{\url{https://github.com/UCSD-AI4H/Medical-Dialogue-System}}
\footnotetext[3]{\url{https://github.com/zhangsheng93/cMedQA2}}
\footnotetext[4]{\url{https://github.com/king-yyf/CMeKG_tools}}
\footnotetext[5]{\url{https://medmcqa.github.io}}
\footnotetext[6]{\url{https://github.com/OpenLMLab/MOSS}}
\footnotetext[7]{\url{https://github.com/Instruction-Tuning-with-GPT-4/GPT-4-LLM}}

\subsection{Real-world Dialogue Records}
We choose two public datasets collected from online forums as the sources of real-world dialgoue records, namely, MedDialog~\citep{chen2020meddiag} and cMedQA2~\citep{cMedQA2-2018}. MedDialog contains over 3 million multi-turn conversations between doctors and patients and the topic focuses on medical consultation scenarios. cMedQA2 contains 108k single-turn conversations encompassing both consultation and advisory situations. After filtering records using keyword filters and named entity recognition, We randomly select 400k and 20k samples from each dataset respectively as source samples for SFT dataset construction.




Real-world dialogues are noisy in linguistic patterns and expression styles of different doctors vary; moreover, doctor's responses might not align with the identify of an AI doctor. In order to obtain high quality conversation samples, we employ the language ability of general LLMs to reconstruct the entire dialogue. We design some prompts (Figure~\ref{fig:prompt1}) for GPT-3.5, following several rules stated below. 

\begin{itemize}
    \item Remove colloquial expressions, address inconsistencies in the doctor's language use, and distill more uniform expressions from the LLM.
    \item Adhere to the key information in the original doctor's response, and based on that, provide an appropriate explanation and supplement to the original answer, rephrasing it in a more detailed and logical manner.
    \item Rewrite or remove responses that shouldn't be made by an AI doctor, such as viewing imaging materials or asking the patient to register for an appointment.
\end{itemize}


Figure \ref{fig:adaption} displays a sample in resulted dataset. After reconstruction, the doctor's response aligns with the identity of the AI medical assistant. The overall response adheres to the key information provided in the original answer by the doctor, and offers more comprehensive assistance to the patient with richer content. 

\begin{figure*}
    \centering
    \includegraphics[width=16cm]{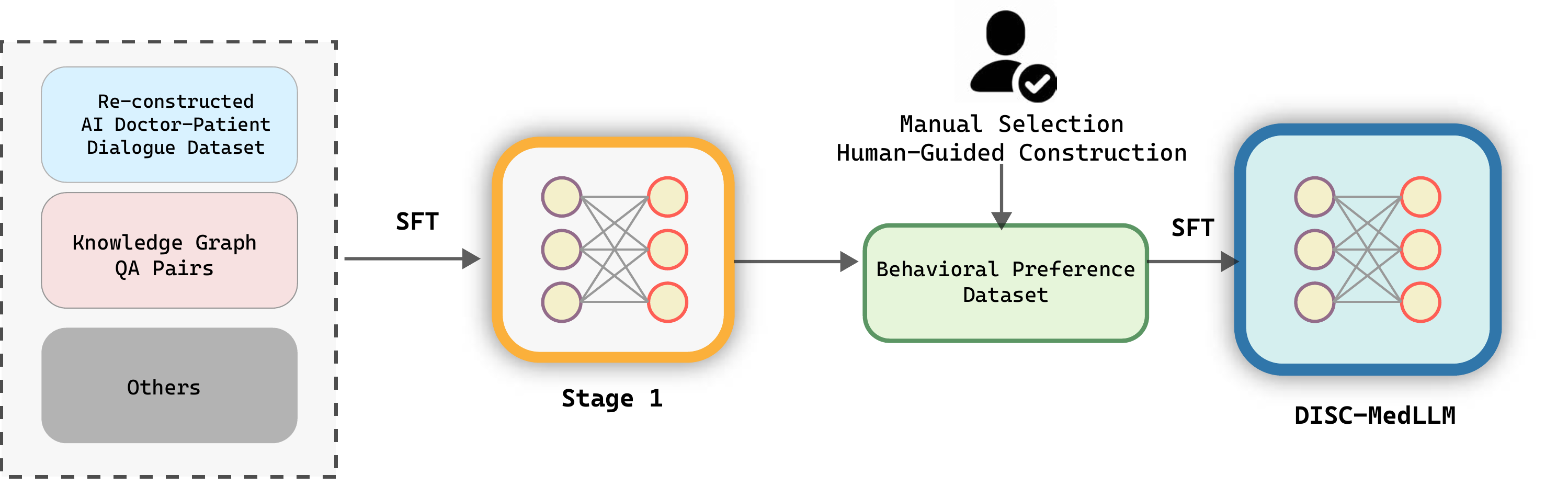}
    \caption{The two stage training process of DISC-MedLLM. Stage 1 equips the model with l with domain knowledge and medical dialogue
capability using diverse datasets. In Stage 2, the model's performance is enhanced through the Behavioral Preference Dataset, which aligned with human preferences..}
    \label{fig:figure3}
\end{figure*}

\subsection{Knowledge Graph QA pairs}
The medical knowledge graph contains a large amount of well organized medical expertise. Generating QA training samples based on it can enable us to obtain data with relatively low noise compared with real world samples. We construct QA pairs based on a Chinese medical knowledge graph which contains over 10k diseases, nearly 20k medications, and more than 10k symptoms. Centered on diseases, we sample the knowledge graph based on the department information of the disease nodes, following the department distribution in the original MedDialog data. We obtain QA pairs through two steps.

\begin{itemize}
    \item \textbf{Step 1:} Leveraging the powerful capabilities of GPT-3.5, we first transform the sampled knowledge into simple natural language QA pairs. The knowledge information about diseases is provided to GPT-3.5 and is converted into a natural language representation in the format <instruction, knowledge>.

    \item \textbf{Step 2:} Based on these simple QA pairs, GPT-3.5 transforms them into single-turn dialogues in a medical scenario, enhancing their diversity and the richness of the language expression.
\end{itemize}
Over 50k diverse medical scenario conversation samples have been generated in total. The details of the prompts used to generate conversations refer to Figure \ref{fig:prompt2} and Figure \ref{fig:prompt3}.

\subsection{Human Preferences Guided Conversation Samples}

To further enhance the model's performance and align its response and behavior more closely with human preferences, we need to utilize a higher-quality dataset that aligns more closely with human preferences for secondary supervised fine-tuning in the final training stage. We manually select approximately 2,000 high-quality, diverse samples suitable for adaptation from the MedDialog and cMedQA2 datasets that were not used in the previous data construction process.

Depending on the sample's consultation scenario, disease severity, and other variables, we select several examples to be reconstructed by GPT-4 and then manually revise them to align more closely with our preferences for AI doctor behavior and response quality. Subsequently, using a few-shot approach, we provid these examples to GPT-3.5 to generate 2,000 high-quality behavior-tuning samples under human supervision.

\subsection{Others}

\paragraph{MedMCQA} ~ {MedMCQA~\cite{pal2022medmcqa} is a multiple-choice Q\&A dataset in the medical field in English, and it provides expert-annotated explanations for each question. We utilize it to generate professional medical QA samples to enhance the model's expertise in Q\&A capabilities. We refine the questions and correct answers from the multiple-choice questions using GPT-3.5, combined with explanations to generate QA format samples, and then translate them into Chinese. Using this approach, we construct about 8k samples, of which approximately 2k samples retain the multiple-choice format and are directly translated into Chinese.}

\paragraph{General} ~ {We introduce some generic data to enrich the diversity of our training set, mitigating the risk of foundational capability degradation in the model during the SFT training phase, especially if data predominantly leans towards the medical sector~\cite{wen2023chathome}. Specifically, we draw from two general domain supervised fine-tuning datasets: moss-sft-003~\cite{sun2023moss} and alpaca\_gpt4\_data\_zh~\cite{peng2023instruction}. From moss-sft-003, we extract data from the \textit{Brainstorming}, \textit{Role Playing}, and \textit{Harmless categories}, selecting 33k samples at random. For alpaca\_gpt4\_data\_zh, considering it is only used in the final stages of training, where sample volume is reduced, we randomly sample 1k instances.}



\section{Training Details}

We develop our model on top of the Baichuan-13B-Base model, which is an open source LLM with over 13.2 billion parameters that was trained on 1.4 trillion tokens corpus, exhibiting ideal performance in both English and Chinese. As illustrated in the Figure \ref{fig:figure3}, our training is divided into two SFT stages, both of which are completed on 4*A800 GPUs. 

\begin{itemize}
    \item \textbf{Stage 1:} We initially use large-scale samples to imbue the model with domain knowledge and medical dialogue capabilities, including a 420k adapted AI doctor-patient dialogue dataset, 50k knowledge graph constructed QA pairs, MedMCQA, and moss-sft-003. The hyperparameters setting for this training process are as follows: global batch size of 24, learning rate of $1e-5$ with AdamW optimizer, 1 epochs, maximum sequence length of 2048 tokens, warm up steps of 1800 and with no weight decay.
    \item \textbf{Stage 2:} In this stage, we train the model to align with human preferences in terms of behavioral patterns, usefulness, etc., enabling it to perform better in medical dialogue scenarios. We employ a 2k meticulously crafted preference-aligned behavioral preference dataset and combine it with 1k alpaca\_gpt4\_data\_zh data for training. The hyperparameters setting for this training process are as follows: global batch size of 8, learning rate of $5e-6$ with AdamW optimizer, 1 epochs, maximum sequence length of 2048 tokens, with no weight decay.
\end{itemize}

\section{Evaluation Setup}

We evaluate the performance of medical LLMs in two settings, namely, single-turn question answering and multi-turn conversation.

\subsection{Single-turn Question Answering Evaluation}

To evaluate the single-round QA capability of LLMs in providing accurate answers to healthcare related questions, we utilize construct a benchmark dataset including multiple-choice questions based on several public datasets. Although open-ended QA setting has been widely employed to test the interactive capability of LLMs in terms of rule-based metrics (ROUGE, BLUE, etc.), it is suitable for evaluating medical consultation systems which care more about the accuracy instead of free-style generation. We use multiple-choice questions to evaluate different systems and use accuracy as the metric. 

\subsubsection{Multiple-choice Datasets}

We sample cases from two public datasets to construct our evaluation benchmark. \textbf{(1) MLEC-QA}~\citep{li-etal-2021-mlec} is collected from the National Medical Licensing Examination in China (NMLEC). It is divided into five categories: Clinic, Stomatology, Public Health, Traditional Chinese Medicine, and Integrated Traditional Chinese and Western Medicine. We randomly sample 10\% from its test set, resulting in a total of 1,362 questions for evaluation.  \textbf{(2) NEEP} is a collection of multiple-choice questions from the Western Medicine 306 of the National Entrance Examination for Postgraduate (NEEP) that we manually collated. For Western Medicine 306, we acquire questions from the years 2019 to 2021 and use a combined total of 270 questions from 2020 and 2021 for our tests. The overall statistics of the benchmark is shown in Table~\ref{tab:benchmarkDatasets}.

We experiment using both zero-shot and few-shot methodologies. For the few-shot samples, the MLEC-QA examples are chosen from its validation set, while those for NEEP are derived from the 2019 set of questions.




\begin{table}
\centering
\scalebox{1.0}{
\begin{tabular}{@{}ccc@{}}
\toprule
Dataset & \begin{tabular}[c]{@{}c@{}}Test Set\\ Original Size\end{tabular} & Sample Size \\ \midrule
MLEC-QA Clinic & 3362 & 336  \\
MLEC-QA CWM & 2674 & 268  \\
MLEC-QA PublicHealth & 1853 & 185 \\
MLEC-QA Stomatology & 2644 & 264 \\
MLEC-QA TCM & 3086 & 309  \\
NEEP 306 & 270 & 270  \\ \bottomrule
\end{tabular}
}
\caption{Benchmark Evaluation Dataset Details. 10\% of cases are sampled from original datasets to form the benchmark dataset. MLEC-QA has 5 subsets, including Clinic, Traditional Chinese Medicine Combined with Western Medicine, Public Health, Stomatology and Traditional Chinese Medicine. NEEP contains Western Medicine 306.}
\label{tab:benchmarkDatasets}
\end{table}

\begin{table*}[]
\scalebox{0.9}{
\begin{tabular}{ccccccccc}
\hline
\specialrule{0em}{2pt}{2pt}
Method                     & Model               & \begin{tabular}[c]{@{}c@{}}MLEC-QA\\ Clinic\end{tabular} & \begin{tabular}[c]{@{}c@{}}MLEC-QA\\ CWM\end{tabular} & \begin{tabular}[c]{@{}c@{}}MLEC-QA\\ PublicHealth\end{tabular} & \begin{tabular}[c]{@{}c@{}}MLEC-QA\\ Stomatology\end{tabular} & \begin{tabular}[c]{@{}c@{}}MLEC-QA\\ TCM\end{tabular} & NEEP 306       & Avarage        \\ \hline
\specialrule{0em}{2pt}{2pt}
\multirow{4}{*}{\textbf{few-shot}}  & GPT-3.5             & \textbf{58.63}                                           & \textbf{45.9}                                         & \textbf{53.51}                                                 & \textbf{51.52}                                                & \textbf{43.47}                                        & \textbf{44.81} & \textbf{49.64} \\
                           & Baichuan-13b-Chat & 31.25                                                    & 37.69                                                 & 28.65                                                          & 27.27                                                         & 29.77                                                 & 24.81          & 29.91          \\
                           & HuatuoGPT(13B)         & 31.85                                                    & 25                                                    & 32.43                                                          & 32.95                                                         & 26.54                                                 & 24.44          & 28.87          \\
                           & DISC-MedLLM         & \underline{44.64}                                              & \underline{41.42}                                           & \underline{41.62}                                                    & \underline{38.26}                                                   & \underline{39.48}                                           & \underline{33.33}    & \underline{39.79}    \\ \specialrule{0em}{2pt}{2pt}\cline{2-9} \specialrule{0em}{2pt}{2pt}
\multirow{4}{*}{\textbf{zero-shot}} & GPT-3.5             & \textbf{47.32}                                           & 33.96                                                 & \textbf{48.11}                                                 & \textbf{39.77}                                                & 38.83                                                 & \textbf{33.33} & \textbf{40.22} \\
                           & Baichuan-13b-Chat & 44.05                                                    & \textbf{43.28}                                        & \underline{39.92}                                                    & 31.06                                                         & \underline{41.42}                                           & \underline{32.22}    & \underline{38.66}    \\
                           & HuatuoGPT(13B)         & 27.38                                                    & 21.64                                                 & 25.95                                                          & 25.76                                                         & 24.92                                                 & 20.37          & 24.34          \\
                           & DISC-MedLLM         & \underline{44.64}                                              & \underline{37.31}                                           & 35.68                                                          & \underline{34.85}                                                   & \textbf{41.75}                                        & 31.11          & 37.56          \\  \bottomrule
\end{tabular} 
}
\caption{Results of multiple-choice benchmark. The highest score is highlighted in bold, while the second is underscored.}
\label{tab:multi-choice}
\end{table*}

\subsection{Multi-turn Conversation Evaluation}

In the scenario of multi-turn dialogue, it is insufficient to evaluate the system performance using traditional evaluation metrics. While human evaluation may be a preferable solution, it is costly and difficult to reproduce the evaluation across different projects. Given these considerations, we select a small set of dialogue samples based on real-world cases, and propose four metrics focusing on medical conversation. We employ GPT-3.5 play the role of the patient and chat with the model for three rounds. In addition, we utilize external LLMs as the judge.

\subsubsection{Dialogue Evaluation Datasets} 

We choose samples from three public datasets, and manually check the quality of samples. \textbf{(1) CMB-Clin} provides 74 real-world medical cases~\citep{wang2023cmb}, detailing patient history summaries, chief complaints, and various laboratory and imaging tests conducted. We using GPT3.5 to generate a initial question based on the 
 patient's condition for each case. One of the cases wasn't suitable for our evaluation setup, leaving us with 73 cases.\textbf{(2) CMD} is a medical Q\&A dataset with a total of 0.79M Q\&A pairs across six departments, containing consultations with explicit demands and diagnostic queries. We randomly choose 20 questions from each department (internal medicine, surgery, pediatrics, andrology, gynecology, and oncology), and this results in 120 samples in total. \textbf{(3) CMID} is a dataset of user queries in the medical domain, where questions are categorized into symptoms, treatment methods, medications and others~\citep{cmid_chen2020benchmark}. We randomly select 30 samples from each category and result in 120 cases. The final evaluation set contains 313 cases. CMB-Clin simulates real-world consultation process, while CMD and CMID focus on the evaluation from the perspectives of departmental specialities and user intentions.

\subsubsection{Evaluation Metrics for Conversation} In order to perform a systematical evaluation on dialogue capability, we propose four metrics, namely, \textit{proactivity}, \textit{accuracy}, \textit{helpfulness}, and \textit{linguistic quality}.

\begin{itemize}
    \item Proactivity: The doctor can proactively and clearly request the patient to provide more information when the information is insufficient.
    \item Accuracy: The diagnosis or advice provided by the doctor is accurate and has no factual errors. Conclusions are not made arbitrarily.
    \item Helpfulness: The doctor can provide the patient with clear, instructive and practical assistance, to address the patient's concerns.
    \item Linguistic Quality: The doctor correctly understands the patient's query, and the expression of the response is smooth and natural.
\end{itemize}

\subsubsection{GPT4-as-a-Judge} Strong LLM judges like GPT-4 can match both controlled and crowdsourced human preferences well~\citep{zheng2023judging}. In this evaluation, GPT-4 serves as a referee and perform evaluation in providing a rating score from 1 to 5 for each of the four criteria.



\section{Evaluation Results}

We compare DISC-MedLLM with some competitive systems and present results for both single-turn and multi-turn evaluations. 

\subsection{Comparative Models}

We are positioning our model in comparison with three general-purpose LLMs and two specialized conversational Chinese medical LLMs. These include: (1) GPT-3.5~\citep{chatgpt}, one of the most powerful and most widely used LLM developed by OpenAI; (2)GPT-4~\cite{openai2023gpt4} the subsequent iteration of GPT3.5, exhibits the most advanced overall performance among the existed LLM series. (3) Baichuan-13B-Chat~\citep{Baichuan13B}, the chat version of the 13 billion parameter pre-trained Baichuan-13B-Base model; (4) BianQue-2~\citep{chen2023bianque1}, an open-sourced Chinese medical LLM with 6 billion parameters; and (5) HuatuoGPT-13B~\citep{huatuogpt-2023}, a Chinese large language model fine-tuned on both distilled and real-world data for medical use. 
GPT-4 and BianQue-2 have not been extensively tested in multiple-choice question answering, due to billing constraints and non-conformance to the expected output, respectively.

\begin{table*}
\centering
\begin{tabular}{@{}cccccc@{}}
\toprule
\textbf{Model}             & \textbf{Proactivity} & \textbf{Accuracy} & \textbf{Helpfulness} & \textbf{Linguistic Quality} & \textbf{Average} \\ \specialrule{0em}{2pt}{2pt} \midrule
GPT-3.5            & 4.30                 & 4.53              & 4.55                 & 5.00                       & 4.60             \\
GPT-4              & 4.15                 & 4.70              & 4.75                 & 4.96                       & 4.64             \\
Baichuan-13b-Chat & 4.30                 & 4.58              & 4.73                 & 4.95                       & 4.64             \\
BianQue-2         & 3.97                 & 4.36              & 4.37                 & 4.81                       & 4.38             \\
HuatuoGPT(13B)       & 4.40                 & 4.62              & 4.74                 & 4.96                       & 4.68             \\      \midrule
DISC-MedLLM       & 4.64                 & 4.47              & 4.66                 & 4.99                       & 4.69            \\ \bottomrule
\end{tabular}
\caption{Multi-turn conversation results on CMB-clin. The score in each detailed metric is the average of all samples.}
\label{tab:cmb-clin}
\end{table*}

\begin{figure*}
    \centering
    \includegraphics[width=0.7\textwidth]{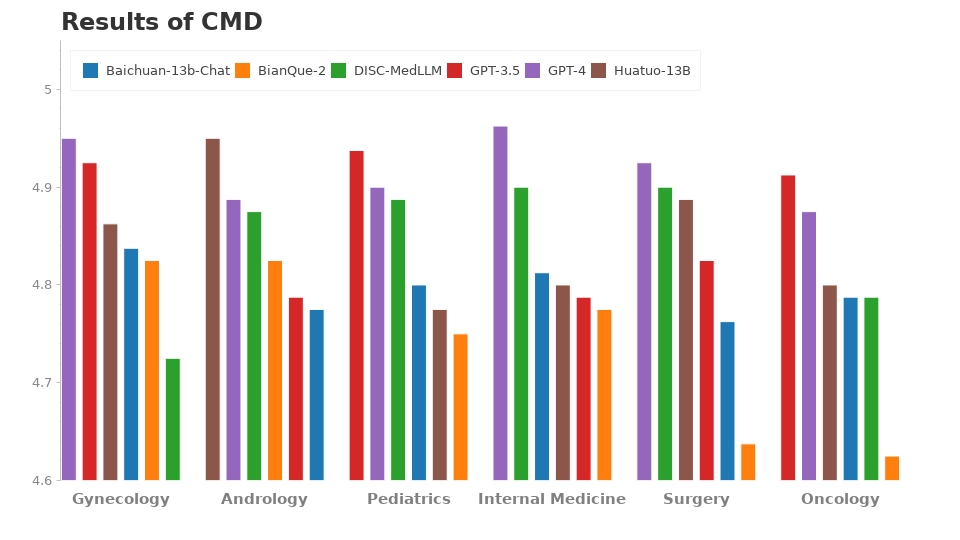}
    \caption{Multi-turn conversation results on CMD. Grouped by different departments and arranged in descending order.}
    \label{fig:figure4}
\end{figure*}

\begin{figure*}
    \centering
    \includegraphics[width=0.7\textwidth]{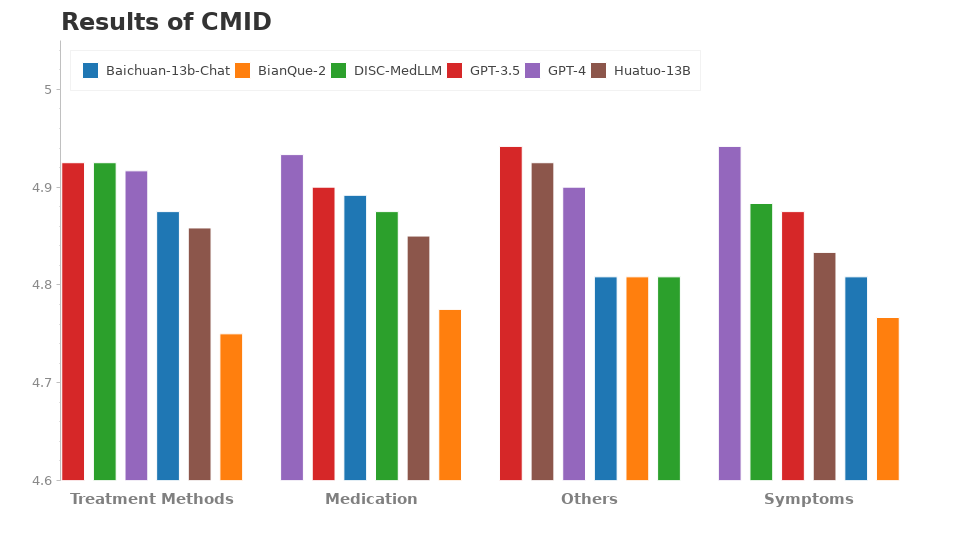}
    \caption{Multi-turn conversation results on CMID. Grouped by different patient intentions and arranged in descending order.}
    \label{fig:figure5}
\end{figure*}

\subsection{Results for Single-Turn QA}

The overall results of multiple-choice assessment are shown in Table \ref{tab:multi-choice}. GPT-3.5 demonstrates an undeniable lead. Our model achieves the second best results in the few-shot setup. In the zero-shot setting, DISC-MedLLM trails behind Baichuan-13B-Chat, ranking the third. It is worthy to know that we perform better than HuatuoGPT(13B) which is trained in reinforcement learning setup.



\subsection{Results for Multi-turn Conversation}


In the CMB-Clin evaluation, DISC-MedLLM garners the top aggregate score, with HuatuoGPT coming in a close second as is shown in Table \ref{tab:cmb-clin}. All three general-purpose models demonstrate commendable performances during this assessment. While GPT-4 excels in the accuracy and usefulness categories, its inclination to address issues within its current interaction leads to a diminished proactivity score compared to GPT-3.5. Significantly, our model registers the highest score in the proactivity criterion, underscoring the effectiveness of our tailored approach for medical model behavioral patterns.

In the CMD sample, as shown in Figure \ref{fig:figure4}, GPT-4 achieves the highest score, followed by GPT-3.5. The medical domain models DISC-MedLLM and HuatuoGPT have identical overall performance scores. When breaking down the scores by six departmental categories, our model outperforms in internal medicine, surgery, and pediatrics, while HuatuoGPT excels in the other three departments.

The situation in CMID is similar to CMD, as is demonstrated in Figure \ref{fig:figure5}, with GPT-4 and GPT-3.5 maintaining the lead. Excluding the GPT family, our model performs the best. It outperforms HuatuoGPT in three intent categories: symptoms, treatment plans, and medications.

The inconsistency in the performance of various models between CMB-Clin and CMD/CMID may primarily result from the different data distributions across the three datasets. CMD and CMID contain more samples with clear questions, where patients describe their symptoms while typically expressing a clear need. The versatile models GPT-3.5 and GPT-4, which excel in multiple aspects, are evidently more adept at handling such situations.




\subsection{Discussion}
While generative models have achieved remarkable improvements in usability for medical interactions, encompassing aspects such as linguistic fluency, semantic comprehension, and the relevance of recommendations, accuracy remains an unyielding concern. Particularly in the medical sphere, disseminating incorrect or deceptive information can lead to graver ethical and practical implications than in other sectors. At present, there's a distinct gap in robust methodologies that can bolster the precision of LLMs within medical contexts. Using retrieval engines to augment LLM responses is a potential avenue; however, the challenges of curating a comprehensive document repository and ensuring alignment between the retriever and query semantics stand as significant barriers. The quest to refine the accuracy of LLMs in healthcare is a pressing challenge that beckons deeper investigation.

\section{Related Works}







In recent years, in order to improve the quality of medical services, reduce medical costs and the unequal distribution of healthcare resources, there has been an increasing focus on the development of intelligent medical dialogue systems~\citep{wang2021pre}. The goal is to facilitate professional, accessible and affordable healthcare resources and help to improve the communication efficiency between healthcare providers and patients. With the advancements in deep learning technologies, researchers have explored various stages of healthcare dialogue systems, including medical entity recognition~\citep{cheng2022named}, symptom identification~\citep{zhang2020mie}, terminology standardization~\citep{zhang2021cblue}, intent classification~\citep{chen2023benchmark}, medical report generation~\citep{gu2020automatic}, dialogue state tracking~\citep{liu2022prompt}, automated diagnosis~\cite{liao2020task,zhong2022hierarchical}, drug recommendation~\citep{garg2021drug}, among other potentially valuable tasks. Although such systems have shown promising results in experimental settings, the challenges associated with the limited scope of applicability and difficulties in integration have hindered their practical application and deployment. 




The relentless advancement of Large Language Models (LLMs), especially like GPT-3.5, ChatGPT~\citep{chatgpt}, GPT-4~\cite{openai2023gpt4}, featuring hundreds of billions of parameters, has unlocked the potential to create highly end-to-end instruction-followed conversational systems~\cite{zhao2023survey}. Distilling the internal knowledge~\cite{gou2021knowledge,chen2022contextual} from these models has emerged as a prominent approach for fine-tuning moderately-sized pre-trained models (like billions of parameters), notably through techniques like self-instruction~\cite{wang2022self}. 

Researchers have made significant progress in developing LLM specifically designed for medical healthcare by distillation from models like ChatGPT~\cite{chatgpt}, as well as domain knowledge from various sources. These models include ChatDoctor, a medical LLM fine-tuned on the LLaMA model using patient-doctor dialogues as training data~\citep{li2023chatdoctor}. Baize-healthcare is another medical adaptation of the Baize chat model, trained on 100k medical dialogs generated by letting ChatGPT chat with itself~\citep{xu2023baize}. MedAlpaca combines Stanford Alpaca and AlpacaLoRA, delivering enhanced models for medical question-answering and dialogue~\citep{han2023medalpaca}. PMC-LLaMA utilizes medical papers to fine-tune LLaMA, aiming to improve medical task performance~\citep{wu2023pmcllama}. Developed by Google, Med-PaLM 2 exhibits great potential for clinical use with impressive performance on various benchmarks~\citep{singhal2023expertlevel}. These advancements demonstrate the growing utility of LLM for medical healthcare.



In terms of medical LLMs in Chinese, we have also witnessed several encouraging outcomes. ~\citet{wang2023huatuo} have constructed knowledge-based instruction data and trained BenTsao (original name: HuaTuo) by adopting LLaMA-7B model as the base model. DoctorGLM~\citep{xiong2023doctorglm} has been demonstrated as an example of fine-tuning LLMs for healthcare purpose with relatively affordable costs. MedicalGPT~\cite{MedicalGPT} is a specialized Chinese medical model. Built upon several foundational models, it underwent incremental pre-training, supervised fine-tuning, and reinforcement learning training. ChatMed~\cite{zhu2023ChatMed} is another Chinese medical LLM. It leverages questions from online consultation websites and distills answers from ChatGPT. The model is fine-tuned using LoRA based on LLaMA-7B. It demonstrates commendable performance in single-turn QA scenarios. HuatuoGPT~\citep{huatuogpt-2023} blends real-world conversations with distilled data acquired from ChatGPT to improve its ability in Chinese healthcare applications.


The distinctive aspect of DISC-MedLLM in comparison to existing medical LLM lies in the knowledge distilled from ChatGPT. To mitigate potential hallucination issues, we carefully construct our data, particularly the labels, leveraging existing medical NLP datasets. Our primary focus is on learning the behavioral patterns and human preferences of ChatGPT, rather than relying on distilling its medical knowledge.



\section{Conclusion}


In this paper, we propose DISC-MedLLM, a comprehensive solution that bridges the gap between general large language models (LLMs) and real-world medical consultation. Our approach leverages ChatGPT to rephrase existing medical NLP datasets to provide accurate and truthful medical responses in end-to-end conversational healthcare services. Through the construction of high-quality Supervised Fine-Tuning (SFT) datasets using strategies such as medical knowledge-graphs, real-world dialogue reconstruction, and human-guided preference rephrasing, DISC-MedLLM surpasses existing medical LLMs in both single-turn and multi-turn consultation scenarios. Experimental evaluations demonstrate its effectiveness in multiple-choice Q\&A and systematic medical consultations. We release the constructed dataset and model weights to further contribute to research and development. We plan to introduce retrieval enhanced DISC-MedLLM in the future, hoping to incorporate additional medical expertise to enhance the model's ability to handle complex and rare medical cases.


\bibliography{Technical_Report}

\appendix
\newpage
\onecolumn
\section{Appendix}
\section{Appendix A: Supplementary Figures}
\setcounter{table}{0}
\setcounter{figure}{0}
\renewcommand{\thetable}{A\arabic{table}}
\renewcommand{\thefigure}{A\arabic{figure}}

\begin{figure*}[h]
    \centering
    \includegraphics[width=16cm]{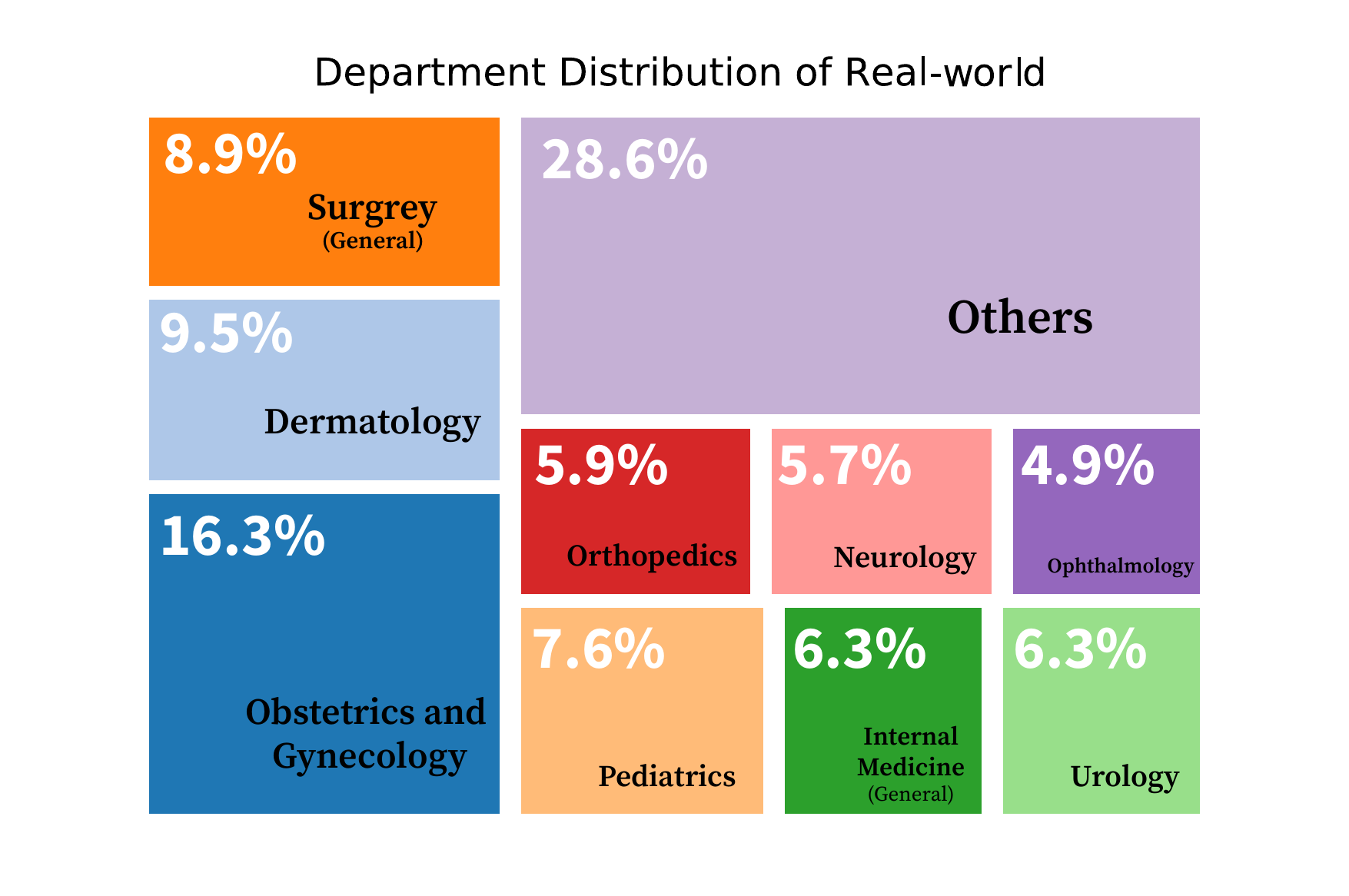}
    \caption{The Distribution of Department in Real-world Dialogue Records. Here exhibits the top 9 departments with the highest proportions, while the rest are categorized as "Others". See the full version in Figure \ref{fig:depa_dist_all}.}
    \label{fig:depa_dist}
\end{figure*}

\begin{figure*}
    \centering
    \includegraphics[width=16cm]{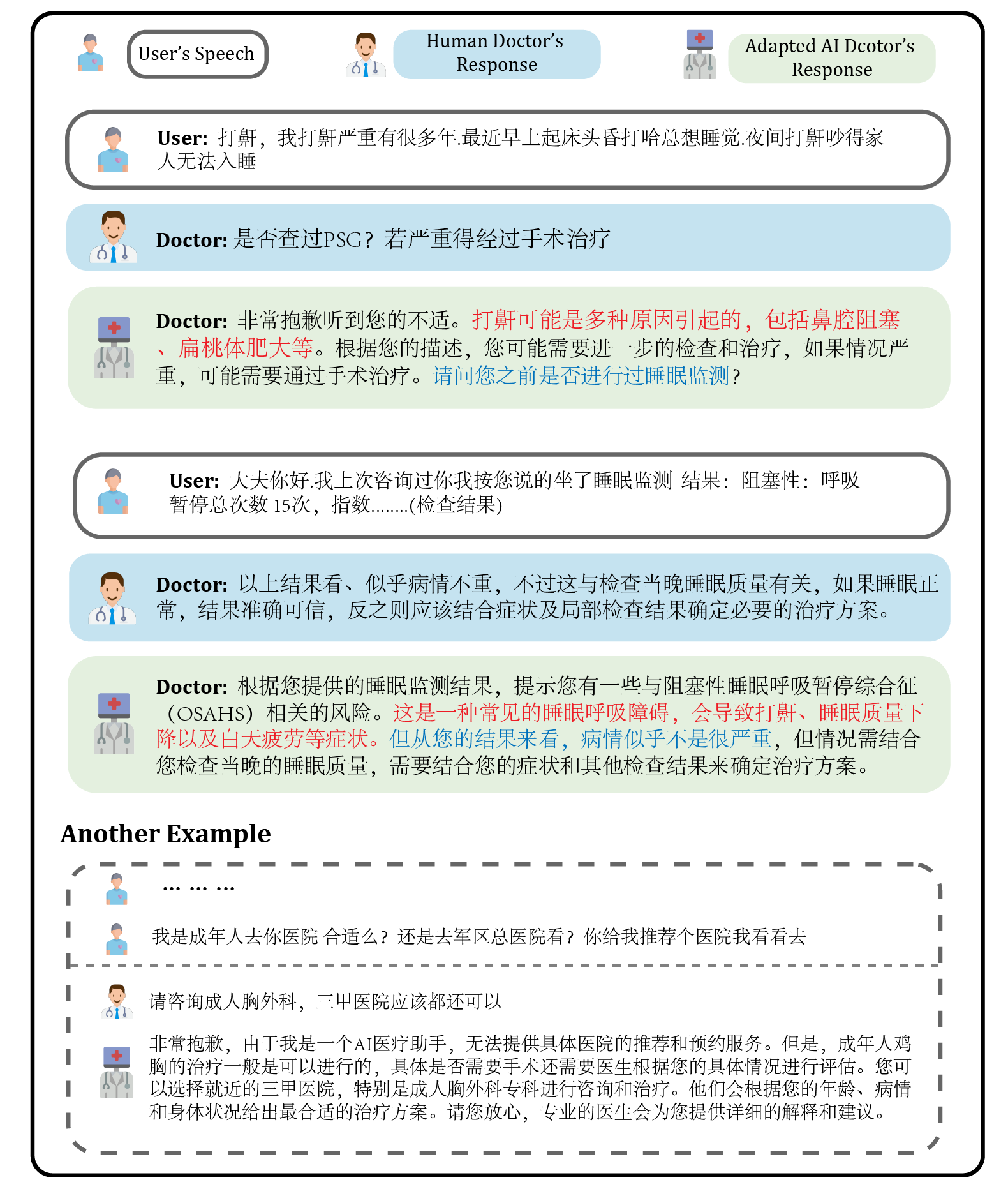}
    \caption{Sample of conversation adaption. The red text indicates knowledge or behavioral patterns sourced from real-world dialogue records, whereas the text in blue denotes content introduced during the re-construction.}
    \label{fig:adaption}
\end{figure*}

\begin{figure*}
    \centering
    \includegraphics[width=16cm]{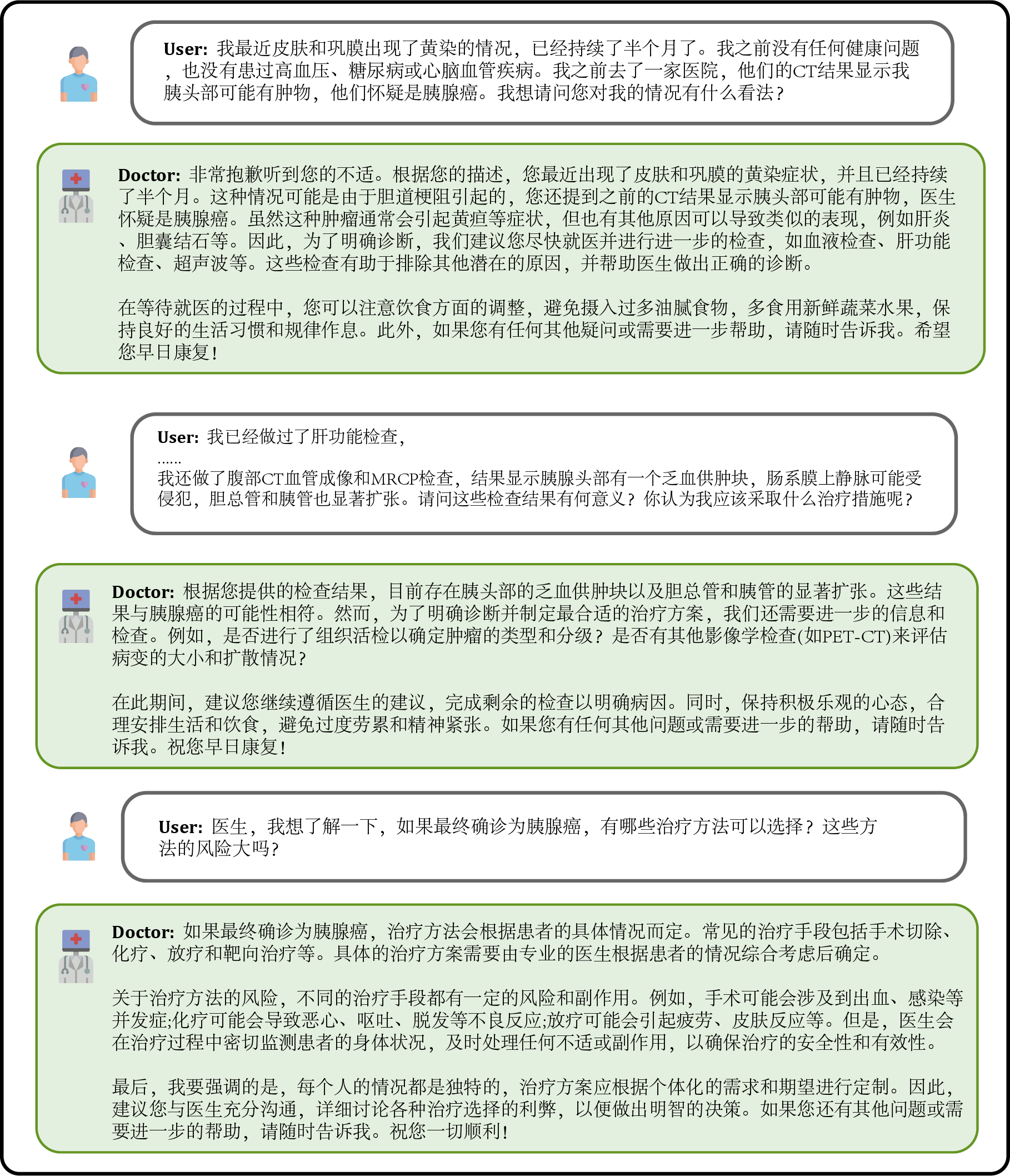}
    \caption{Dialogue Sample of DISC-MedLLM}
    \label{fig:sample}
\end{figure*}

\begin{figure*}
    \centering
    \includegraphics[width=16cm]{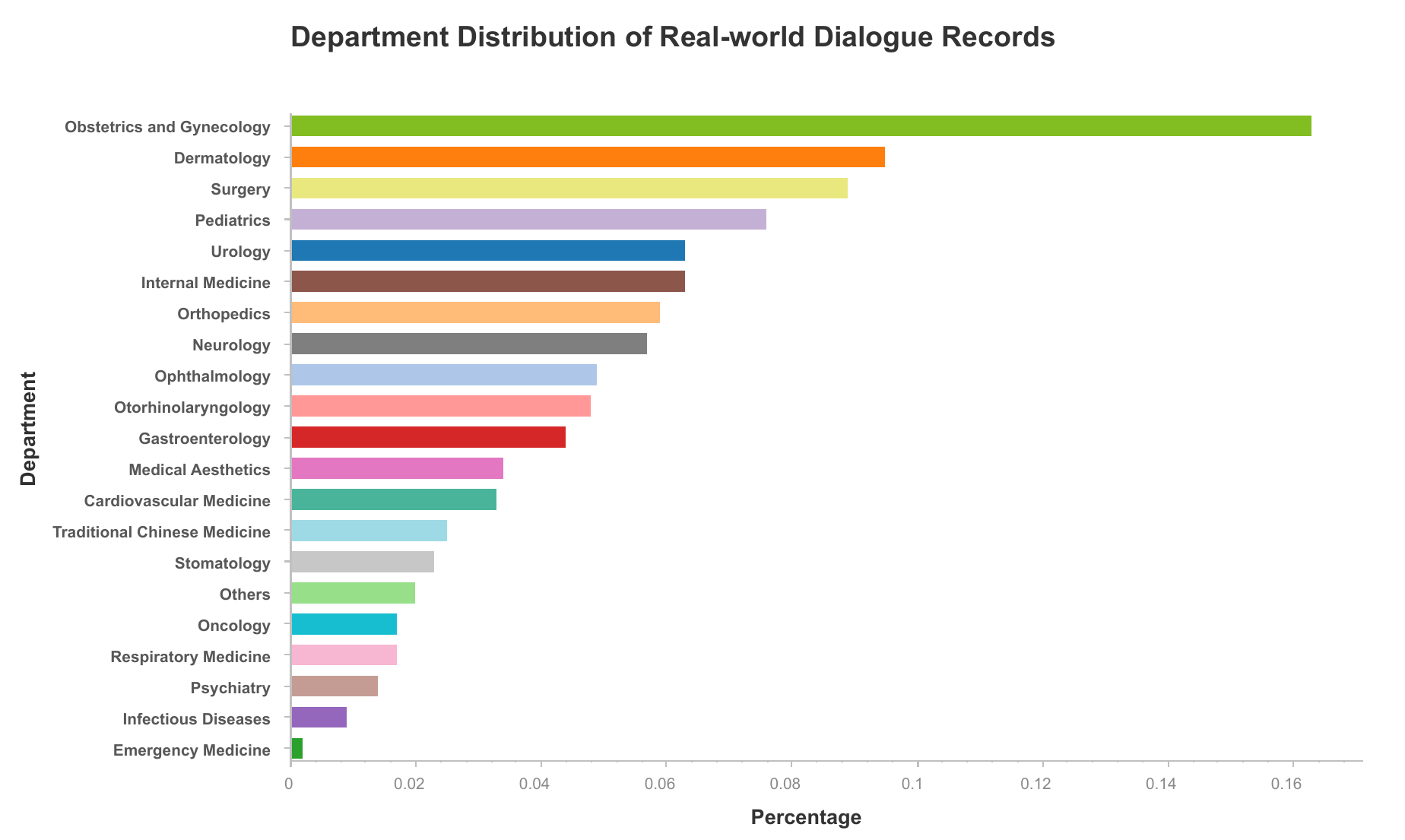}
    \caption{The Distribution of Department in Real-world Dialogue Records. Departments may have hierarchies and we tally based on the most specific category in the list. A record in 'Respiratory Medicine' is counted there, not under 'Internal Medicine'.}
    \label{fig:depa_dist_all}
\end{figure*}
\newpage
\newpage
\newpage
\newpage
\newpage
\newpage
\clearpage
\section{Appendix B: Prompts using in our practices}
\setcounter{table}{0}
\setcounter{figure}{0}
\renewcommand{\thetable}{B\arabic{table}}
\renewcommand{\thefigure}{B\arabic{figure}}
\begin{figure*}[h]
    \centering
    \includegraphics[width=16cm]{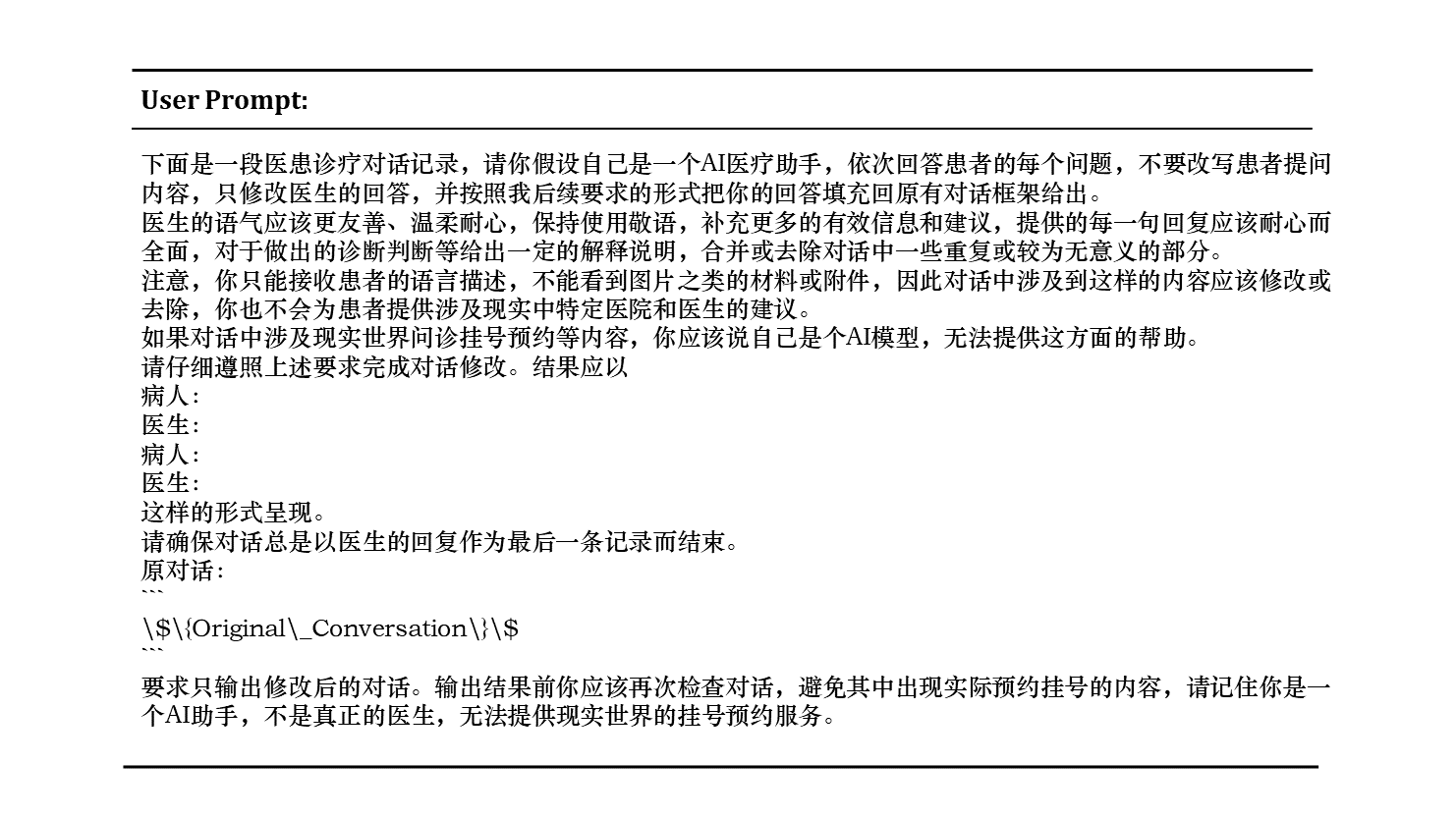}
    \caption{Prompt for Re-constructing Real-world Conversations}
    \label{fig:prompt1}
\end{figure*}

\begin{figure*}
    \centering
    \includegraphics[width=16cm]{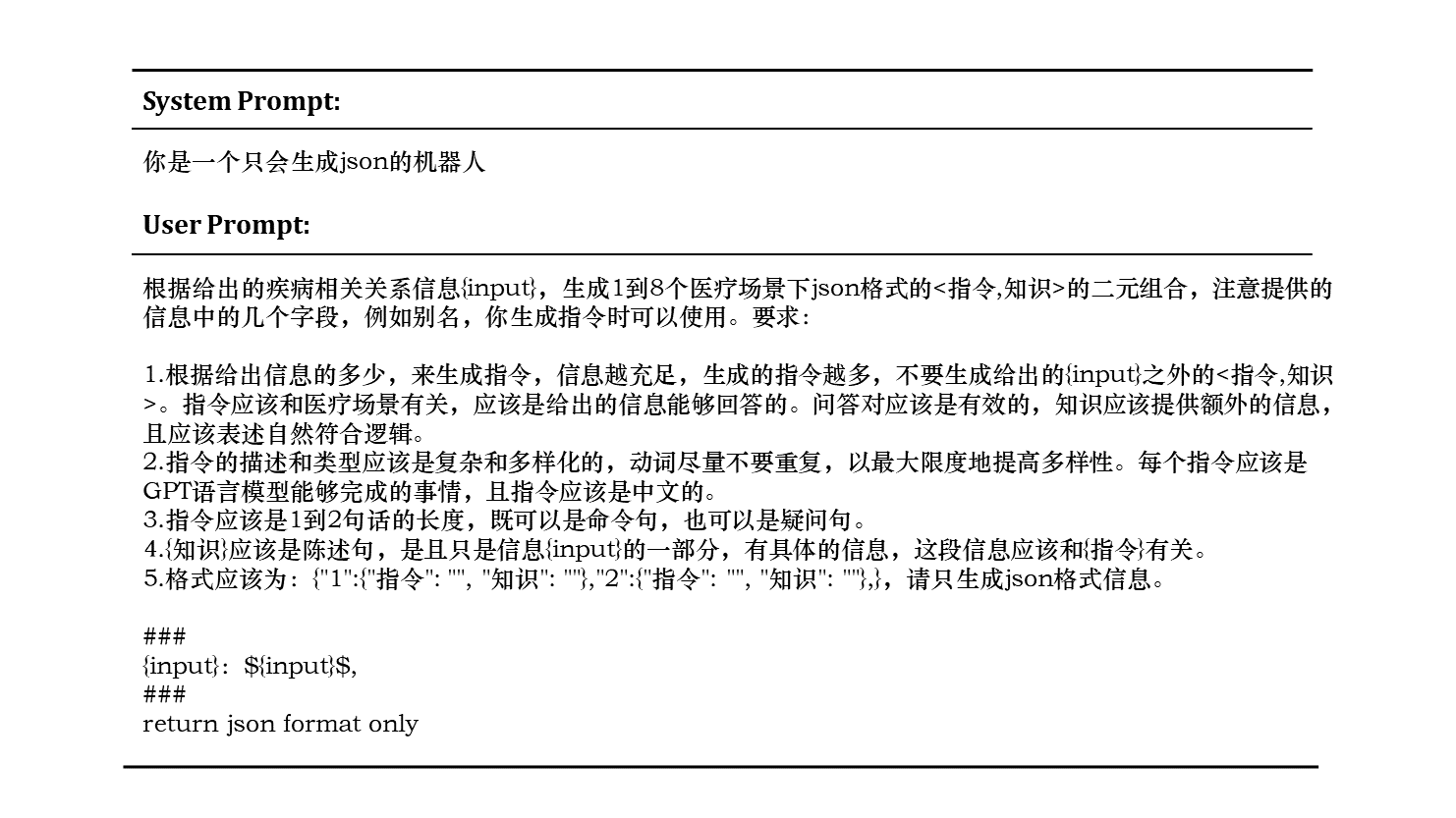}
    \caption{Prompt for Generating QA pairs from Knowledge Graph, Step 1}
    \label{fig:prompt2}
\end{figure*}

\begin{figure*}
    \centering
    \includegraphics[width=16cm]{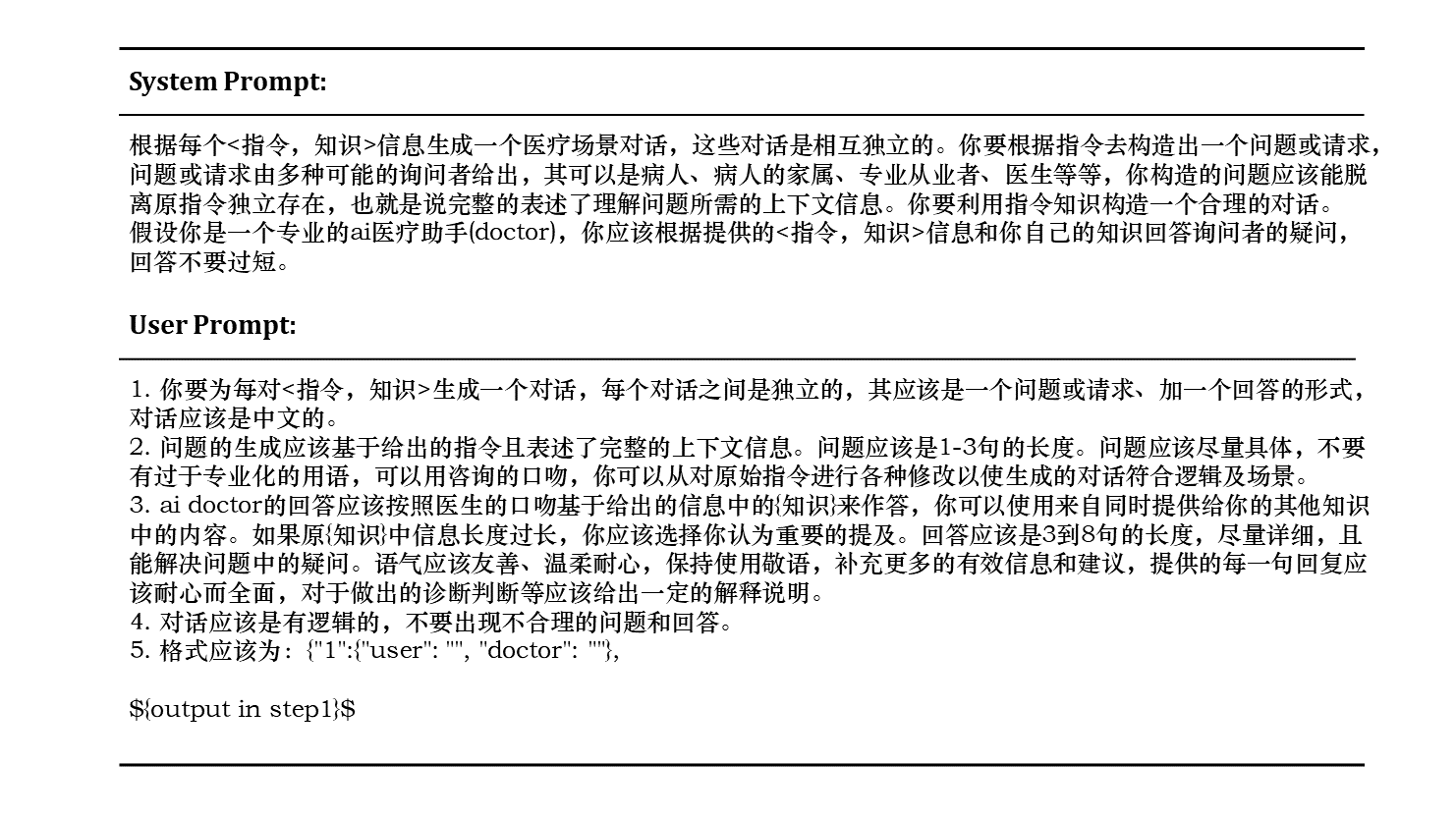}
    \caption{Prompt for Generating QA pairs from Knowledge Graph, Step 2}
    \label{fig:prompt3}
\end{figure*}

\begin{figure*}
    \centering
    \includegraphics[width=16cm]{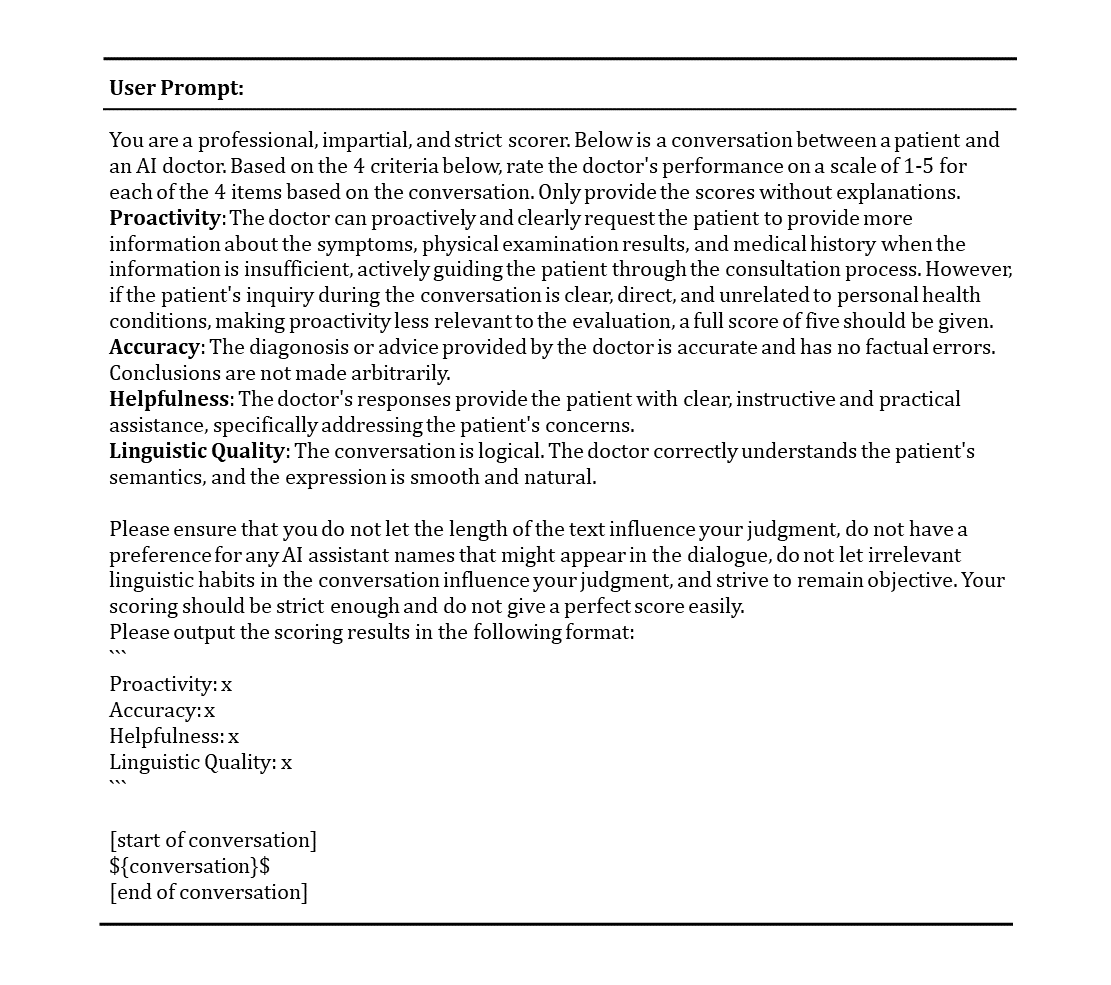}
    \caption{Prompt used in GPT-4-as-a-judge evaluation for multi-turn conversation.}
    \label{fig:prompt4}
\end{figure*}

\end{document}